\documentclass[runningheads]{llncs}

\usepackage{eccv}

\usepackage{algorithm}
\usepackage{algpseudocode}
\usepackage{lipsum}
\usepackage{array}
\usepackage{listings}

\definecolor{softviolet}{rgb}{0.7, 0.4, 0.8}
\definecolor{softblue}{rgb}{0.4, 0.6, 0.8}
\definecolor{forestgreen}{rgb}{0.13, 0.55, 0.13}
\definecolor{forestred}{rgb}{0.55, 0.13, 0.13}
\definecolor{lightgreen}{HTML}{C4F4C8}
\definecolor{lightblue}{HTML}{C4F4E0}
\definecolor{lightgrey}{HTML}{A4A4A4}
\definecolor{ForestGreen}{HTML}{228B22}

\usepackage{eccvabbrv}

\usepackage{graphicx}
\usepackage{booktabs}

\usepackage[accsupp]{axessibility}  

\usepackage[breaklinks,colorlinks,citecolor=eccvblue]{hyperref}

\usepackage{orcidlink} 
  
\begin{document}
\lstset{
    basicstyle=\ttfamily\small, 
    breaklines=true, 
    xleftmargin=0pt, 
    showstringspaces=false 
}

\title{ComiCap: A VLMs pipeline for dense captioning of Comic Panels}

\titlerunning{ComiCap}

\author{Emanuele~Vivoli\inst{1,2}\orcidlink{0000-0002-9971-8738} \and
Niccolò~Biondi\inst{2}\orcidlink{0000-0003-1153-1651} \and
Marco~Bertini\inst{2}\orcidlink{0000-0002-1364-218X} \and
Dimosthenis~Karatzas\inst{1}\orcidlink{0000-0001-8762-4454}}

\authorrunning{E.~Vivoli et al.}

\institute{Computer Vision Center, UAB, Spain \and
MICC, University of Florence, Italy
\\
\email{\{evivoli,dimos\}@cvc.uab.es}\\
\email{\{name.surname\}@unifi.it}}

\maketitle

\begin{abstract}
The comic domain is rapidly advancing with the development of single- and multi-page analysis and synthesis models. Recent benchmarks and datasets have been introduced to support and assess models’ capabilities in tasks such as detection (panels, characters, text), linking (character re-identification and speaker identification), and analysis of comic elements (e.g., dialog transcription). However, to provide a comprehensive understanding of the storyline, a model must not only extract elements but also understand their relationships and generate highly informative captions. In this work, we propose a pipeline that leverages Vision-Language Models (VLMs) to obtain dense, grounded captions. To construct our pipeline, we introduce an attribute-retaining metric that assesses whether all important attributes are identified in the caption. Additionally, we created a densely annotated test set to fairly evaluate open-source VLMs and select the best captioning model according to our metric. Our pipeline generates dense captions with bounding boxes that are quantitatively and qualitatively superior to those produced by specifically trained models, without requiring any additional training. Using this pipeline, we annotated over 2 million panels across 13,000 books, which will be available on the project page \url{https://github.com/emanuelevivoli/ComiCap}.
\keywords{Dense Caption \and Comics panels \and Vision Language Models}
\end{abstract}

\begin{figure}[ht]
    \centering
    \includegraphics[width=\linewidth]{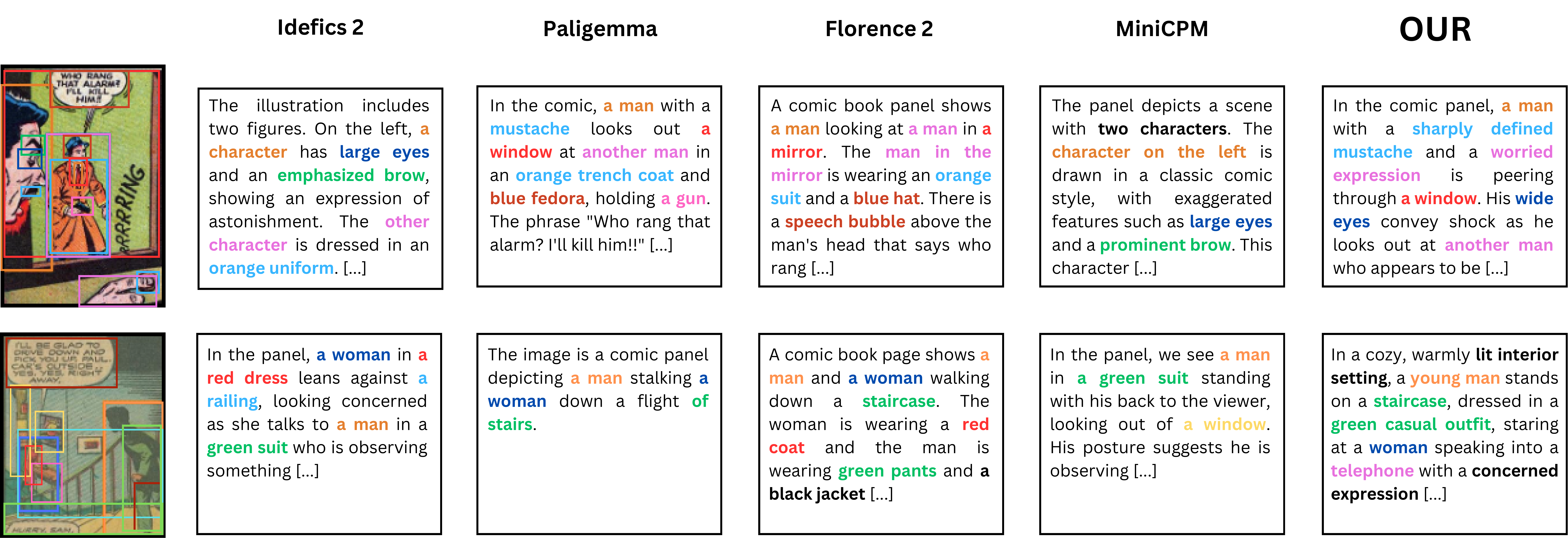}
    \caption{VLMs dense captions compared to our pipeline.}
    \label{fig:enter-label}
\end{figure}

\section{Introduction}
\label{sec:intro}

Comics represent a highly complex medium for computational analysis, yet they are easily understood by humans—except for individuals within the Blind or Low Vision community. Recent studies \cite{li_manga109dialog_2023,ramaprasad_comics_2023,sachdeva2024manga} have addressed this gap by developing dialog generation tasks to assist People with Visual Impairments (PVI). These tasks aim to transcribe all spoken text, sorted by appearance, and associate it with the corresponding character’s name. To support these efforts and facilitate new methods, several benchmarks have been introduced \cite{vivoli_comics_2024,vivoli_comix_2024} that handle multiple tasks across various comic styles, from detection to dialog generation. However, when consuming a comic book solely through transcribed dialog, a crucial element is missing: context. Context, defined as a description of the scene’s happenings, can be extrapolated using recent Vision-Language Models (VLMs) through captioning \cite{rigaud2024toward}. A comprehensive model to assist PVI should generate comic dialog transcriptions while providing dense captions for specific panels and descriptions of characters when necessary.

Despite rapid advancements in dialog transcription \cite{sachdeva2024manga,vivoli_comix_2024,sachdeva_tails_2024}, no research has adequately addressed the task of panel and character description, particularly through a grounded approach that includes bounding boxes associated with generated captions to enhance explainability.

In this paper, we focus on generating dense captions for comic panels using existing VLMs without additional training. Our contributions are as follows:
\begin{itemize}
\item We propose a two-stage metric based on automatic key-element extraction and BERT-score evaluation to assess the presence of important attributes in VLM-provided captions.
\item We annotate 1.5k panels with captions and attribute lists, benchmarking existing open-source VLMs for comic panel caption generation using the proposed metric.
\item We identify the top-performing Vision-Language model and propose a multi-stage pipeline that refines panel captions from fine- to coarse-grained, demonstrating superior dense captions compared to specifically trained models.
\end{itemize}

With the proposed pipeline, we annotate more than 2 million panels, providing dense captions to the research community.

The rest of the paper is structured as follows: in Section \ref{sec:related}, we provide an overview of existing tasks in the comic domain, as well as an overview of visual language models. In Section \ref{sec:cap-attr}, we detail our attribute extraction process starting from VLM captions. In Section \ref{sec:metric}, we justify and describe the proposed metric. Section \ref{sec:pipeline} evaluates VLMs against the benchmark dataset using our metric and iteratively designs our pipeline. Finally, in Section \ref{sec:conclusion}, we present qualitative results and discuss future work that our approach enables.

\section{Related Work}
\label{sec:related}

\paragraph{\textbf{Comics tasks.}}
In the field of comics, common tasks span from detection, segmentation, and element linking (such as speaker identification) to clustering (character re-identification). Recently, some works have focused on designing dialog generation tasks as a proof of concept for single-page comics analysis, starting from simpler atomic tasks \cite{sachdeva2024manga} and later also extending to include names \cite{sachdeva_tails_2024}. However, many of these tasks lack proper evaluation, leading recent works to provide metrics and benchmarks for detection, linking, and dialog generation tasks \cite{vivoli_comics_2024,vivoli_comix_2024}. Despite these advances, a story cannot be fully understood, especially by those who cannot easily access comic images, when relying solely on dialog generation and character naming. Numerous events occur within a scene and across the gutters \cite{iyyer_amaz_2017}. A natural progression is the detailed panel description through captioning to provide context for every significant scene \cite{rigaud2024toward}. Recent VLMs have demonstrated surprising performance in generating both short and long captions, particularly in out-of-domain settings.

\paragraph{\textbf{Vision-Language models.}}
Recently, vision and language models have garnered significant attention. A common practice involves integrating a vision encoder into large language models (LLMs), employing various approaches \cite{minigpt4, flamingo, llava, llavanext}, effectively giving LLMs “eyes.” This simple approach can be extended with multiple modifications to accept interleaved (image-text) data \cite{laurençon2024mattersbuildingvisionlanguagemodels}, high-resolution images \cite{llavanext, minicpmhq}, and generate text output \cite{minigpt4, llava, MiniCPM-V}, or even bounding boxes \cite{beyer2024paligemmaversatile3bvlm, xiao_florence2_2023} and segments, using VQ-GAN mask quantized vector indices \cite{beyer2024paligemmaversatile3bvlm} or actual segment perimeter indices \cite{xiao_florence2_2023}. These models are typically trained for a broad range of tasks, following (i) pretraining (only training the mapping operation from vision space to language space), (ii) finetuning (unfreezing the language decoder and, in some cases, also the vision encoder), and (iii) instruct fine-tuning/downstream task setting (unfreezing everything, or training with LoRA). This method slowly adapts the model to align image and text representations in the same space, enabling it to process and understand images and, in some cases, provide detection boxes that validate the model’s understanding, thereby aiding human reasoning about the detected elements.

\paragraph{\textbf{Dense captioning.}} Recent advancements in image captioning, such as those by Vinyals et al. \cite{vinyals_show_2015} and Anderson et al. \cite{anderson_bottom_2018}, have laid the foundation for dense captioning, an advanced extension of traditional image captioning. Dense captioning \cite{johnson_densecap_2015, krishna_dense_2017} involves generating detailed descriptions for multiple regions within an image, integrating object detection with captioning \cite{sermanet2014overfeatintegratedrecognitionlocalization, girshick2014richfeaturehierarchiesaccurate, ren2016fasterrcnnrealtimeobject}. This method significantly enhances visual content description over conventional single-sentence captions. Dense captioning has evolved to include video applications \cite{krishna2016visualgenomeconnectinglanguage} and has been incorporated into vision-language multi-task models like Florence2 \cite{xiao_florence2_2023}, which are trained on extensive datasets such as FLR-5B. Despite these advancements, dense captioning remains challenging in certain media like comics, which are underrepresented in training datasets for vision-language models (VLMs).

\section{Captioning and Attribute Extraction}
\label{sec:cap-attr}
\begin{figure}[t]
\centering
\includegraphics[width=0.9\linewidth]{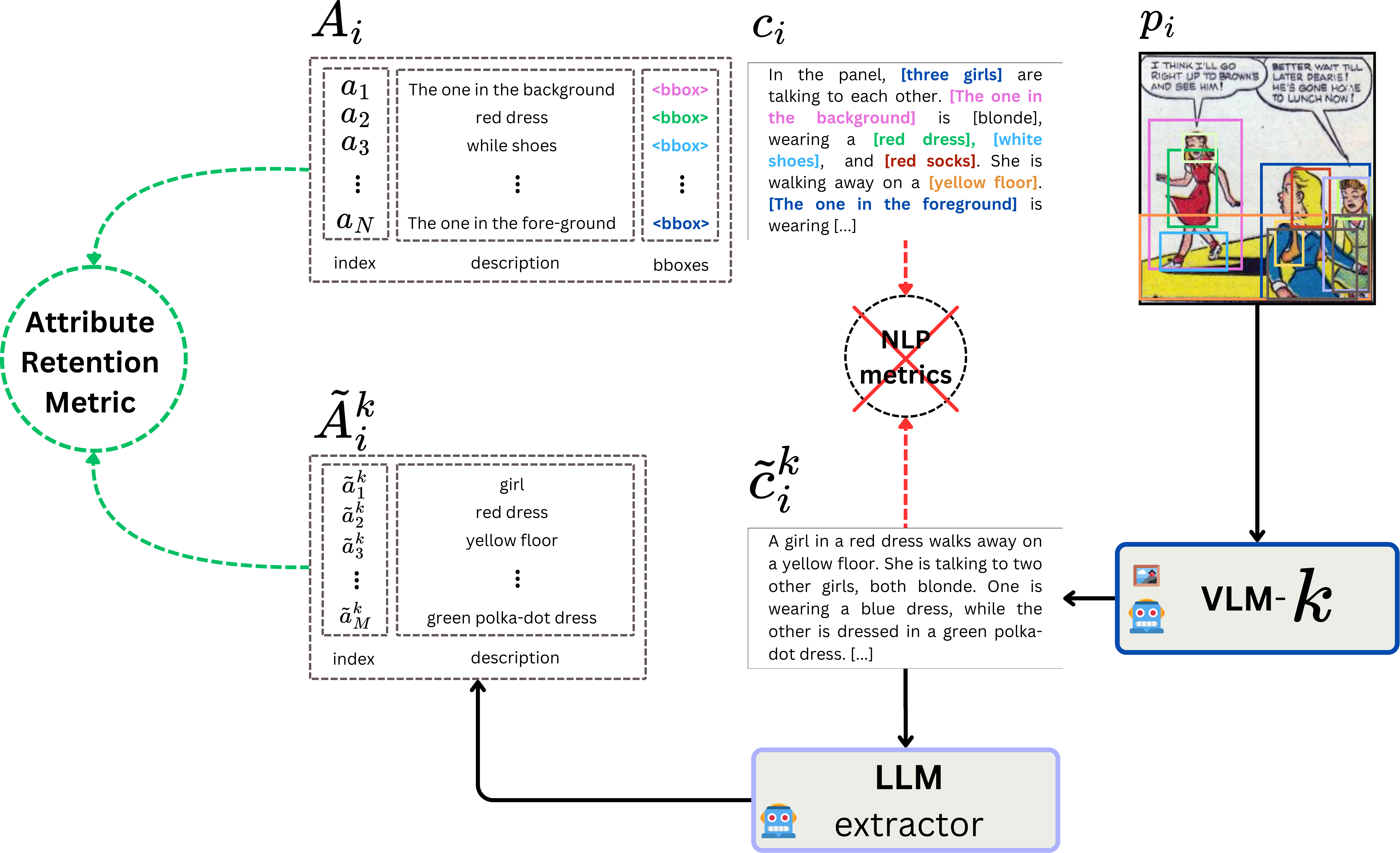}
\caption{Attribute extraction from VLMs captions.}
\label{fig:attribute-extraction}
\end{figure}

Our goal is to generate a dense caption from a comic book panel that includes all important attributes related to the scene and characters, along with their bounding boxes. In this section, we describe how we utilize VLMs in a zero-shot manner to generate a list of attributes that can later be grounded (see Figure \ref{fig:attribute-extraction}). These attributes are used to evaluate the caption's quality against a designed test set of panel captions and attributes.

\paragraph{\textbf{Captioning models.}}
Among the recent VL models, the majority have been trained for captioning \cite{caffagni2024revolutionmultimodallargelanguage}.

Idefics2 \cite{laurençon2024mattersbuildingvisionlanguagemodels}, a fully autoregressive architecture with 8 billion parameters, leverages the SigLIP-SO400M vision encoder \cite{zhai2023sigmoidlosslanguageimage} for robust visual processing and the Mistral-7B language model \cite{jiang2023mistral7b} for advanced language understanding capabilities. The model is trained using the OBELICS dataset \cite{laurençon2023obelicsopenwebscalefiltered} for foundational multimodal understanding and is further refined via instruction tuning with The Cauldron dataset, which includes a wide range of vision-language tasks.

The MiniCPM-llama3-V-2.5 model \cite{MiniCPM-V} integrates a sophisticated architecture blending elements from SigLIP-400M \cite{zhai2023sigmoidlosslanguageimage} and Llama3-8B-Instruct \cite{dubey2024llama3herdmodels}. It was trained using a unique set of training data that includes CommonCrawl and Code Pretrain datasets, C4, and smaller contributions from sources such as Arxiv and Open Web Math, among others. For handling high-resolution images, the model employs a resampling technique that adjusts the input images to optimal sizes and resolutions without extensive padding or reshaping, maintaining efficiency and minimizing data distortion. Fine-tuning involves reinforcement learning from AI feedback (RLAIF-V), using AI-generated responses to refine the model’s alignment with human-like reasoning, substantially reducing errors such as hallucinations. This model supports a wide array of tasks, including high-quality OCR for large-scale images and complex instruction following.

\paragraph{\textbf{Grounding models.}}
Some vision-language models also support generating bounding boxes associated with textual hooks.

PaliGemma \cite{beyer2024paligemmaversatile3bvlm} is a versatile 3B Vision-Language Model (VLM) that combines the capabilities of a 400M SigLIP vision encoder \cite{zhai2023sigmoidlosslanguageimage} with a 2B Gemma language model \cite{gemmateam2024gemmaopenmodelsbased}. PaliGemma follows a prefix-LM masking paradigm, always prefixing image tokens followed by the prompts/questions, applying usual causal masking to the generated text. The model is designed to deliver bounding boxes and segmentation masks using two additional vocabularies: a 1024-bin vocabulary for x-y detection boxes and a 128 VQ-VAE tokenized single object mask tokens. The training involves multiple stages, starting with individual pretraining of unimodal components, followed by multimodal pretraining, resolution increase, and finally, task-specific transfer training, including captioning.

Florence2 \cite{xiao_florence2_2023} is a foundation model that adopts a unified, prompt-based approach to accommodate a diverse set of vision and vision-language tasks. It integrates a vision encoder and text-location tokenizers, which are provided as input to an encoder-decoder transformer in a sequence-to-sequence framework. The model is trained on the FLD-5B dataset, a meticulously curated collection featuring over 5.4 billion annotations across 126 million images. Florence2 processes multimodal inputs and outputs text and location tokens, enabling it to perform tasks ranging from object detection and image captioning to visual grounding and segmentation. Florence2 performs dense captioning in a two-step process: first, generating a caption of varying lengths, and second, using the caption for text-phrase grounding.

These models were selected for their variability in architecture, training procedure, fine-tuning datasets, and supported tasks. In particular, all these models support image captioning. As illustrated in Figure \ref{fig:attribute-extraction}, given a panel $p_i \in P$, with $P$ being the set of panel images, the first phase consists of generating a caption $\tilde{c}_i$ for the panel $p_i$ with the above models $x^k \in X$, with k being in $\{1, 2, 3, 4\}$. All models are prompted with the text “describe the image in detail.” These provided captions $\tilde{c}_i^{k}$, with longer or shorter text depending on the model’s specifics, are used to extract the attribute list.

\paragraph{\textbf{Attribute extraction.}}
The generated caption should retain all the panel’s objects and attributes (e.g., objects in the scene, important elements, day or night attributes) and characters’ objects and attributes (e.g., clothing, facial and body characteristics, expressions, specific poses) for every significant character in the scene. These are the objects and attributes elements we are interested in.

In the Natural Language Processing field, when comparing  sentences to each other (in our case, raw captions), it has been observed that the METEOR metric most closely aligns with human preferences for multiple-choice captions given an image \cite{banerjee-lavie-2005-meteor}. However, this can be misleading, as our preference is for captions that retain all important elements and attributes of an image, even if this sacrifices some fluency and brevity. Thus, we propose a custom procedure that employs LLMs to perform the task of attribute extraction. Specifically, given the panel image $p_i$ and the caption $\tilde{c}_i^{k}$ generated by the model $x^{k}$, we aim to extract from $\tilde{c}_i^{k}$ a set of attributes $\tilde{A}^{k}_i = \{ \tilde{a}_1^{k}, \tilde{a}_2^{k}, \dots, \tilde{a}_M^{k}\}$ that is most similar to the ground truth attribute set $A_i = \{ a_1, a_2, \dots, a_N \}$. The set of ground truth attributes has been used to generate the ground truth caption $c_i$. As described above, our goal is to obtain similar attribute sets and not necessarily BLEU, ROUGE, or METEOR-optimized captions. Thus, we analyse the attributes set $\tilde{A}_i^{k} $ for all models k, that we obtain employing an LLM specifically prompted for performing this task. The prompt is provided in the project repository. 

\section{Attribute Retaining Metric}
\label{sec:metric}
As illustrated in Figure \ref{fig:attribute-extraction}, we are interested in a metric that can compare two sets of elements: one produced by the chosen model and the other being the ground truth. The metric should be able to associate the elements $a_i, i \in \{1, \dots, N\}$ with the predicted $\tilde{a}_j^{k}, j \in \{1, \dots, M\}$, even if they are not exact string matches. To achieve this, we design a two-step procedure.
First, we compute a pairwise BERT-score among all possible $(i, j)$ pairs. Since the BERT-score is bounded between 0 and 1, we choose a threshold to avoid misleading associations. The threshold $\tau$ is chosen among 0.5 and 0.99, for all models. All the detected associations are then substituted in the predicted attributes set by the respective ground truth elements, resulting in the modified set $\tilde{A}^{*k}_i$.
The second step of the metric involves applying the Jaccard similarity, or intersection-over-union metric, to the sets of elements.

The complete pseudo code is provided in Algorithm \ref{algo:metric}.

\begin{algorithm}
\caption{Attributes Retaining Metric Calculation}
\label{algo:metric}
\begin{algorithmic}[1]
\Procedure{CalculateARM}{\textcolor{softviolet}{$C$}, \textcolor{softblue}{$\tilde{C}^{k}$}, \textcolor{softblue}{$\tau$}}
    \State \textbf{Input:} Ground truth captions \textcolor{softviolet}{$C = \{c_i | i = 1, \ldots, |P|\}$}, predicted captions \textcolor{softblue}{$\tilde{C}^{k} = \{\tilde{c}^{k}_i | i = 1, \ldots, |P|\}$} obtained with model \textcolor{forestgreen}{$x^k$}, vision language model \textcolor{forestgreen}{$x^k$}, threshold \textcolor{softblue}{$\tau$}
    \State \textbf{Output:} Jaccard similarities \textcolor{softviolet}{$J^{k} = \{ J^{k}_i | i= 1, \dots, |P| \}$} with $p_i$ being the panel and $|P|$ being the size of panels set.

    \For{each \textcolor{forestgreen}{$i$} from $1$ to $|P|$}
        \State Extract predicted entity sets \textcolor{softviolet}{$\tilde{A}_i$} from predicted caption \textcolor{softblue}{$\tilde{c}_i$} using model \textcolor{forestgreen}{$x^k$}
        \State Calculate BERT-score \textcolor{softblue}{$BS(\tilde{A}_i, A_i)$} for each \textcolor{softviolet}{$A_i$}
    \EndFor
    
    \For{each \textcolor{forestgreen}{$i$} from $1$ to $|P|$}
        \State Initialize cleaned set \textcolor{softviolet}{$\tilde{A^*}_i \leftarrow \emptyset$}
        \State Initialize cleaned set \textcolor{softblue}{$J_i \leftarrow \emptyset$}

        \For{each element \textcolor{softviolet}{$\tilde{a}$} in \textcolor{softviolet}{$\tilde{A}_i$}}
            \If{\textcolor{softblue}{$BS(\tilde{a}, A_i) \geq \tau$}}
                \State Replace \textcolor{softviolet}{$\tilde{a}$} with the matching element \textcolor{forestgreen}{$a \in A_i$}
                \State Add \textcolor{forestgreen}{$a$} to \textcolor{softviolet}{$\tilde{A^*}_i$}
            \Else
                \State Add \textcolor{softviolet}{$\tilde{a}$} to \textcolor{softviolet}{$\tilde{A^*}_i$}
            \EndIf
        \EndFor
        \State Calculate the Jaccard similarity \textcolor{softblue}{$Jacc(\tilde{A^*}_i, A_i)$}
        \State Save it in  \textcolor{softblue}{$J_i$}
    \EndFor
    \State \textbf{Return:} \textcolor{softblue}{$J_i$} for $i \in 1, \dots, |P|$.

\EndProcedure
\end{algorithmic}
\end{algorithm}

Finally, we compute the predicted set accuracy $ARM$ as the average Jaccard similarity: $ \frac{1}{N} \sum_{i=1}^{N} J_i $.

\section{Pipeline}
\label{sec:pipeline}

\paragraph{\textbf{Ablation.}}
We have a collection of VLMs capable of captioning comic book panels. These model-generated captions are synthesized by an LLM, specifically GPT4o-mini, which extracts the attributes from the panel captions and provides a list of objects and attributes present in the captions. 
The prompt used to extract the attribute set is provided in the project repository. 
Finally, we obtain a Jaccard similarity score from the BERT-score filtered attribute sets, which we use to assess the model’s precision in retaining all important elements in the caption. The results are provided in Table \ref{tab:ablation}.

\begin{table}[ht]
\centering
\caption{Performance Metrics for Different Models}
\label{tab:ablation}
\begin{tabular}{|l|c|c|c|c|}
\hline
\textbf{Model} & \textbf{ROUGE} & \textbf{BLEU} & \textbf{METEOR} & \textbf{ARM (ours)} \\ \hline
PaliGemma & 0.13 $\pm$ 0.02 & 0.01 $\pm$ <0.001 & 0.05 $\pm$ 0.001 & 0.22 $\pm$ 0.11 \\
Idefics2 & 0.19 $\pm$ 0.07 & 0.29 $\pm$ 0.13 & 0.19 $\pm$ 0.08 & 0.23 $\pm$ 0.10 \\
Florence2 & 0.31 $\pm$ 0.12 & 0.17 $\pm$ 0.11 & 0.17 $\pm$ 0.09 & 0.24 $\pm$ 0.11 \\
MiniCPM & 0.38 $\pm$ 0.12 & 0.34 $\pm$ 0.07 & 0.29 $\pm$ 0.12 & 0.36 $\pm$ 0.11 \\ \hline
\end{tabular}
\end{table}

As seen from the table, MiniCPM retains the highest ARM score, which is not in line with other metrics results such as METEOR, BLEU and RIUGE. This is thanks ti the design if our metric, which enphasise the presence of attributes and objects, and not bigram exact co-occurrency. An additional post-processing technique applied to all models captions involves removing OCR-detected text from the caption. MiniCPM has been pre-trained for OCR transcription, which is evident from its captions, thus this techniques might have impacted the results too.

Once the best VLM is chosen, we can explore an additional setting: improving the captions by separating panel and character captions.
Specifically, we employ the DASS model \cite{dass} for detecting character boxes and provide the model with the panel and each of the character cut-out boxes.
The MiniCPM model, as illustrated in the model description, has been trained for instruction following and refined with reinforcement learning techniques, making it suitable for following instructions. 
We leverage this property by providing two different prompts to the model: one for panels and one for characters. Both prompts are structured so that the output includes the caption and the attribute list together. We successfully parse the output into a CSV file and a text caption file using the appropriate prompt. The prompts are provided in the project repository. 

\begin{figure}[t]
\centering
\includegraphics[width=\linewidth]{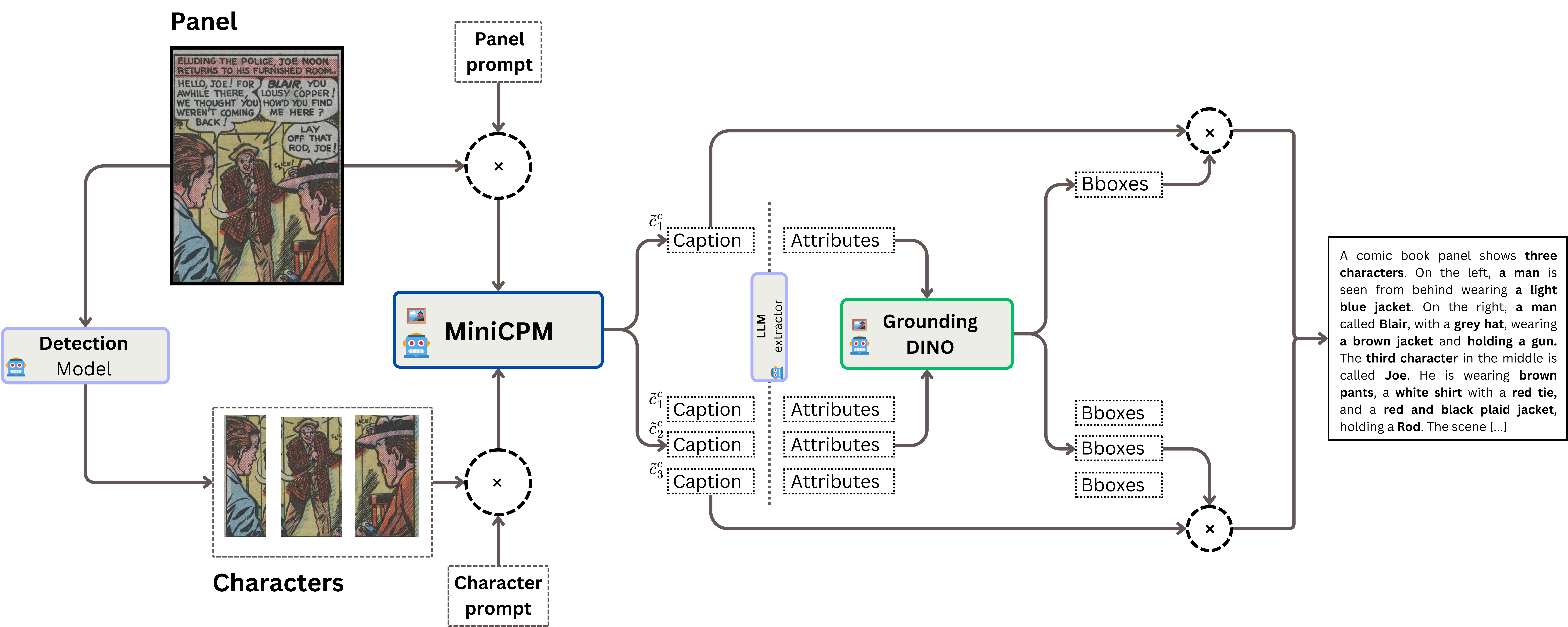}
\caption{Pipeline details for the MiniCPM model using the panel-characters setting.}
\label{fig:pipeline}
\end{figure}

Figure \ref{fig:pipeline} provides an overview of the pipeline, differentiating the two prompts for panels and characters, and obtaining the captions and attribute lists simultaneously. Notably, this instruction-following property is not available in other VLMs.
We empirically discovered that using the panel-characters setting is more favorable compared to using a single caption for the panel.

\paragraph{\textbf{Text Grounding.}}
To achieve dense captioning, we must address the inclusion of bounding boxes. Regardless of the chosen model, an additional VL model can be employed to detect each object in the attributes set. In the case of Florence2, the grounding model is Florence2 itself. However, for MiniCPM, we use GroundingDINO, which has demonstrated significant performance in zero-shot out-of-vocabulary detection, including in contexts like manga \cite{sachdeva2024manga}.

We apply GroundingDINO on the attribute list extracted by MiniCPM. As GroundingDINO sometimes misses objects due to them being unknown, we also ask MiniCPM to provide additional synonym terms to support the attributes in the attribute set.

\begin{figure}[t]
\centering
\includegraphics[width=0.75\linewidth]{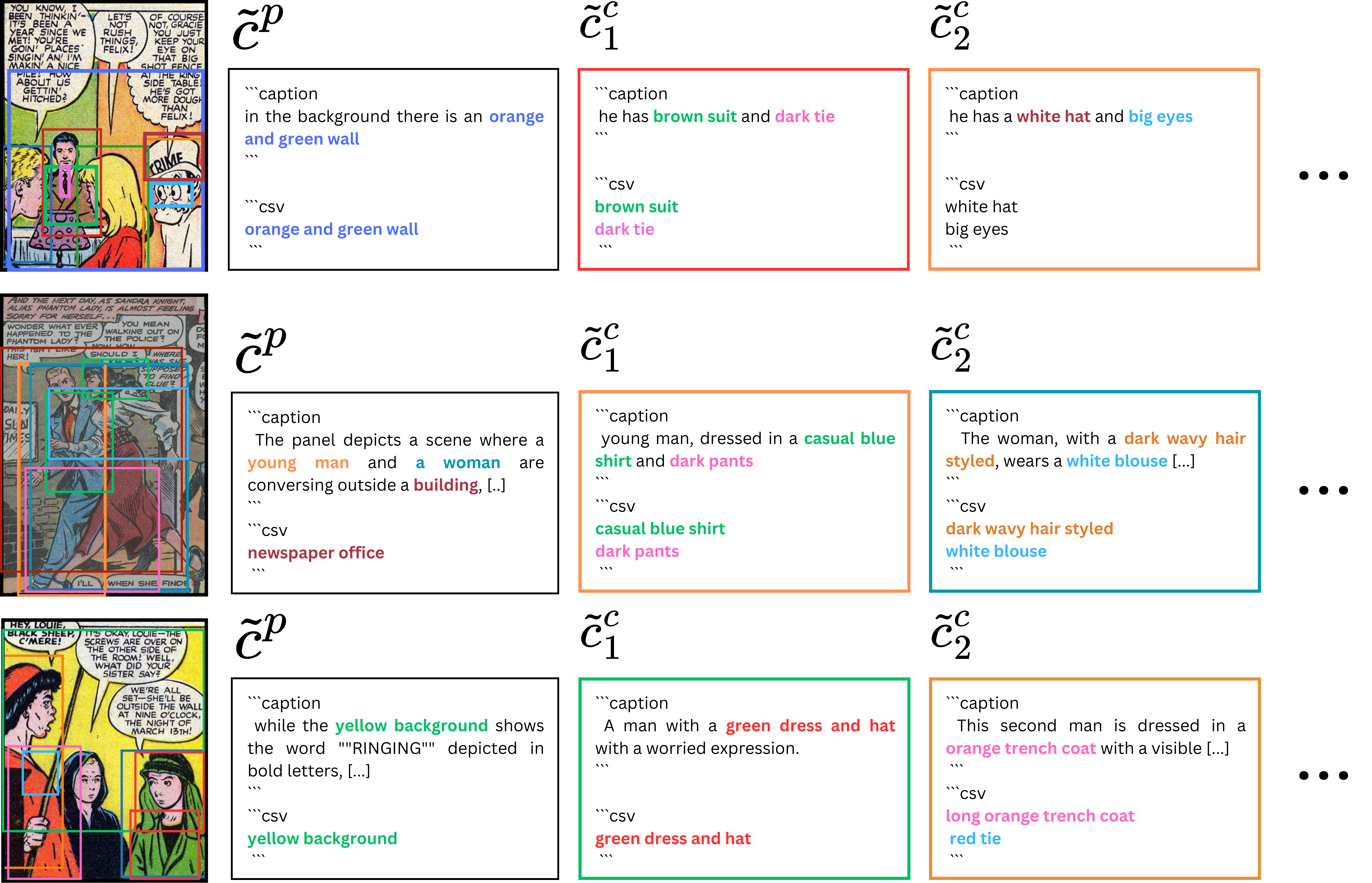}
\caption{Examples of the MiniCPM output, ready to be parsed into attribute lists (csv) and captions (txt).}
\label{fig:parse-minicpm}
\end{figure}

Figure \ref{fig:pipeline} provides an overview of the full pipeline, utilizing MiniCPM with the panel-characters setting and the final version of the prompts. In Figure \ref{fig:parse-minicpm}, we provide examples of MiniCPM outputs, which we parse by searching for the “csv” and “caption” tags.

\begin{figure}[t]
\centering
\includegraphics[width=0.9\linewidth]{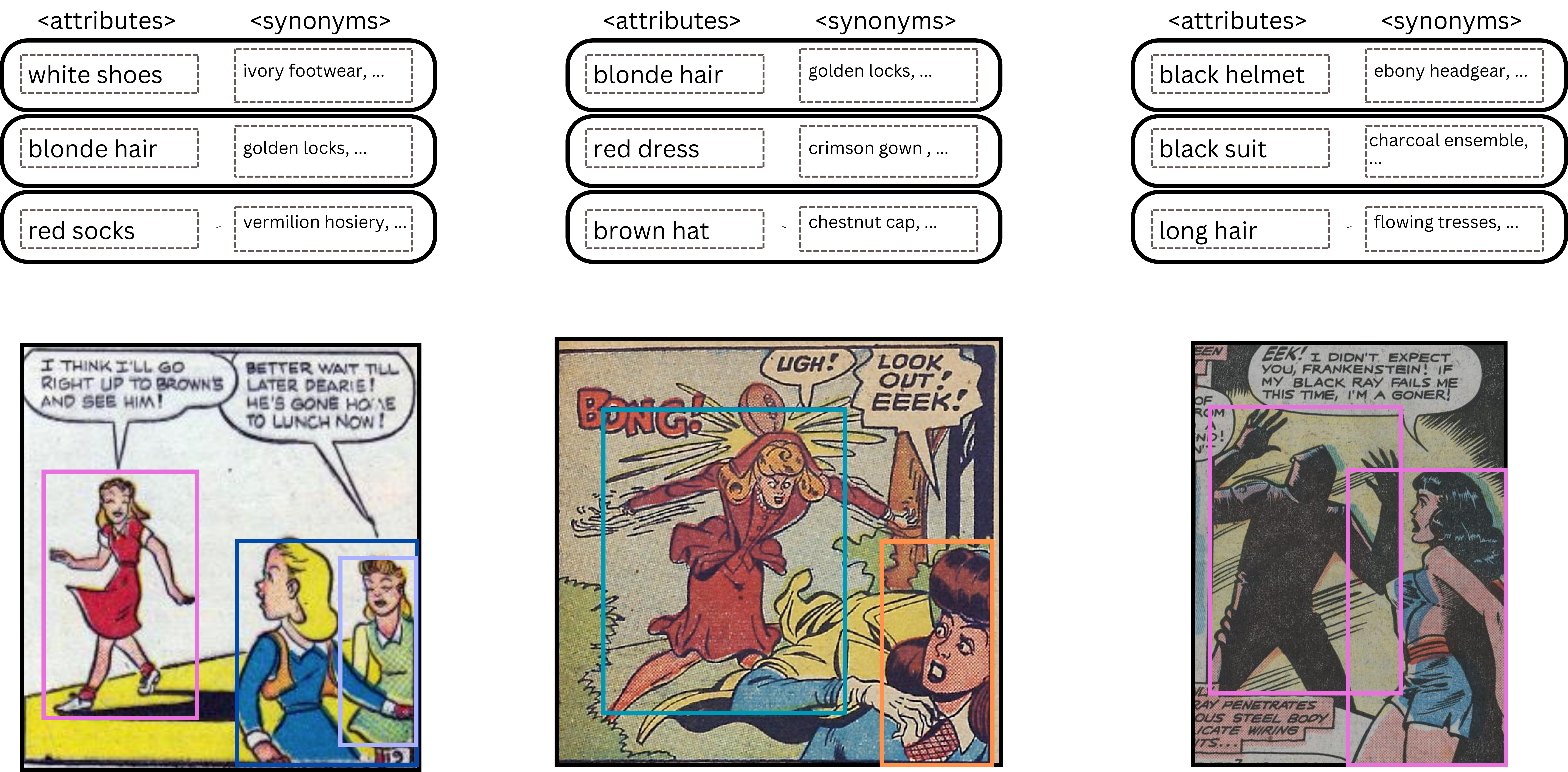}
\caption{Examples of GroundingDINO detections: correctly identified elements (left), missed ones (center), and wrongly detected objects (right).}
\label{fig:gdino-output}
\end{figure}

Examples of GroundingDINO-detected attributes are provided in Figure \ref{fig:gdino-output}.

\paragraph{\textbf{ComiCap Dataset.}}
Copyrights present a significant issue in comics. However, some websites provide sufficiently old, out-of-copyright comic books that are freely accessible and downloadable. We have collected over 13k books from the Digital Comic Museum website, which has been active since 2011 and contains a collection of more than 22k books. Some of these books were unavailable or corrupted, resulting in a final collection of 13k books. Recent work has trained (under limited training sets) and tested various comic style object detection models \cite{vivoli_comics_2024}. The authors demonstrated that, among single- and double-step CNN detectors and transformer models, the best-performing model for comic style is FasterRCNN (for panels) and DASS (for characters). Therefore, we adopted these models to automatically extract panels and characters in the DCM-13k dataset, similar to \cite{iyyer_amaz_2017}. This procedure resulted in over 1.5M panels and 2.06M characters detected, culminating in a densely captioned comics panels dataset, which we provide to the research community. Some qualitative results of the generated captions are shown in Figure \ref{fig:qualitative}.

\begin{figure}[t]
\centering
\includegraphics[width=0.95\linewidth]{imgs/comicap-last.pdf}
\caption{Qualitative results of our pipeline applied to the DCM-13k dataset for the creation of the ComiCap Dataset.}
\label{fig:qualitative}
\end{figure}

\section{Conclusion and Future Work}
\label{sec:conclusion}

In this paper, we have demonstrated an automatic method for generating dense captions of comic panels without requiring additional training, leveraging existing Vision-Language Models (VLMs). We identified limitations in existing captioning metrics and proposed a novel attribute-based metric that combines BERT-score and Jaccard similarity to assess attribute coverage in captions. We benchmarked various VLMs using this metric on a specifically annotated test set. The best-performing model was further explored through prompt engineering to develop a comprehensive dense captioning pipeline, which also incorporates GroundingDINO for text grounding. To the best of our knowledge, this is the first work to generate dense captions for comic panels using a VLMs pipeline without additional training.

Our contributions include the release of the code, the dense captioning test set, and the ComiCap dataset to the research community, aimed at advancing automatic comics analysis for People with Visual Impairments (PVI) and the Blind community.

\paragraph{\textbf{Future Work.}}
There are several avenues for future research. First, improving the accuracy of attribute extraction and text grounding by integrating more sophisticated models or combining multiple VLMs could yield better results. Second, expanding the dataset to include a wider variety of comic styles and languages could enhance the robustness and generalizability of the pipeline. Third, incorporating user feedback, especially from the PVI and Blind community, could help refine the captions to better meet their needs. Finally, exploring real-time captioning systems and their applications in educational and recreational contexts for visually impaired individuals would be a significant step forward.

%
%
\bibliographystyle{splncs04}
\bibliography{ref,ref2}

\end{document}